\documentclass{article}
\usepackage{spconf,amsmath,graphicx,hyperref}
\usepackage{booktabs}
\usepackage{array}
\usepackage{microtype}
\usepackage{float}
\usepackage{balance}
\usepackage[T1]{fontenc}
\usepackage[utf8]{inputenc}
\hypersetup{hypertexnames=false}

\title{Recoverable Geographic Location Information in Earth-Observation Embeddings}
\name{Peiwen Zhang$^{1}$ \quad
Kristie Hu$^{1}$ \quad
Jovana Knezevic$^{2}$ \quad
Shunde Yin$^{1}$ \quad
Kyle Gao$^{3}$
}

\address{$^{1}$University of Waterloo \quad
$^{2}$University of Cambridge \quad
$^{3}$Aalto University}

\begin{document}
\maketitle

\begin{abstract}
Earth-observation (EO) foundation models provide reusable embeddings, yet downstream task accuracy does not reveal whether these representations encode geographic information, which may be beneficial for location-aware applications but potentially detrimental when representations invariant to geographic location are desired. We therefore evaluate the \emph{geographic coordinate robustness} of Tessera v1, Tessera v1.1, and AlphaEarth by testing whether coordinates can be predicted from the embedding representations using 284 quality-verified European solar farms from 2024. We assessed geographic information content information through the association between cosine and geodesic distances and through prediction of projected coordinates in EPSG:3035. Embeddings from all three EO foundation models contain recoverable geographic information. All prediction models significantly outperform training-range uniform random sampling baselines, with AlphaEarth exhibiting the strongest distance association and lowest mean geodesic error. Both Tessera variants also yielded higher geographic distance correlations than the Sentinel-2 controls. These findings motivate geographic information content as an additional criterion for auditing EO foundation models.
\end{abstract}

\begin{keywords}
Representation learning, embeddings, foundation models, spatial representations, remote sensing
\end{keywords}

\section{Introduction}

Earth-observation (EO) foundation models provide reusable pixel-level embeddings for downstream analysis. Tessera summarizes annual Sentinel-1 and Sentinel-2 time series in 128 dimensions \cite{feng2026tessera}, whereas AlphaEarth integrates annual satellite images in a 64-dimensional field at 10~m resolution \cite{brown2025alphaearth}. Their downstream accuracy, however, does not reveal how strongly the embedding representations encode geographic location information, which may be beneficial for location-aware tasks but undesirable when geography-agnostic representations are required. We therefore ask whether geographic locations can be predicted from Tessera v1, Tessera v1.1, and AlphaEarth embeddings within a single semantic class: solar farms.

Geography-aware self-supervised learning, SatCLIP, Tile2Vec, and GeoCLIP all use geographic coordinates or spatial proximity as an explicit training signal---as location-encoder inputs in SatCLIP and GeoCLIP \cite{klemmer2025satclip}, \cite{cepeda2023geoclip}, as a geo-location pretext classification target in geography-aware SSL \cite{ayush2021geography}, and as the basis for positive-pair construction in Tile2Vec \cite{jean2019tile2vec}. These studies do not, however, quantify how much geographic location information (lat, lon) remains recoverable from general-purpose EO embeddings that were produced without explicit coordinate inputs. 
Recent benchmarks emphasize standardized model comparison \cite{dionelis2024evaluating} and broaden the geographic coverage of evaluated datasets \cite{marsocci2024pangaea}. Spatial generalization also depends on the relationship between training and test data: performance can vary with their spatial separation \cite{rolf2021generalizable} or with geographic distribution shifts \cite{koh2021wilds}. Random cross-validation can also overstate spatial transfer when nearby observations enter both training and test sets \cite{roberts2017crossvalidation}, \cite{ploton2020spatial}, \cite{karasiak2022spatial}, \cite{kattenborn2022spatially}.  To ensure a consistent comparison while accounting for spatial dependence, we compare all three embedding representations using the same solar farms sites and spatial folds. Because spatially related farm pairs are not independent \cite{guillot2013dismantling}, we resample at the level of spatial dependency groups rather than treating the 40,186 pairs as independent observations.

Our contribution is a controlled assessment of embedding geometry and coordinate sensitivity:
\begin{itemize}
    \item \textbf{Geographic distance association:} We test whether cosine distance is associate with geodesic distance using embeddings of a controlled surface, namely photovoltaic surfaces of solar farms.
    \item \textbf{Coordinate sensitivity:} We evaluate embedding robustness to geographic coordinates using spatially held-out coordinate probes against appearance-based and random baselines.
\end{itemize}

\section{Data and Methodology}

\begin{figure}[htpb]
\centering
\includegraphics[width=0.95\linewidth]{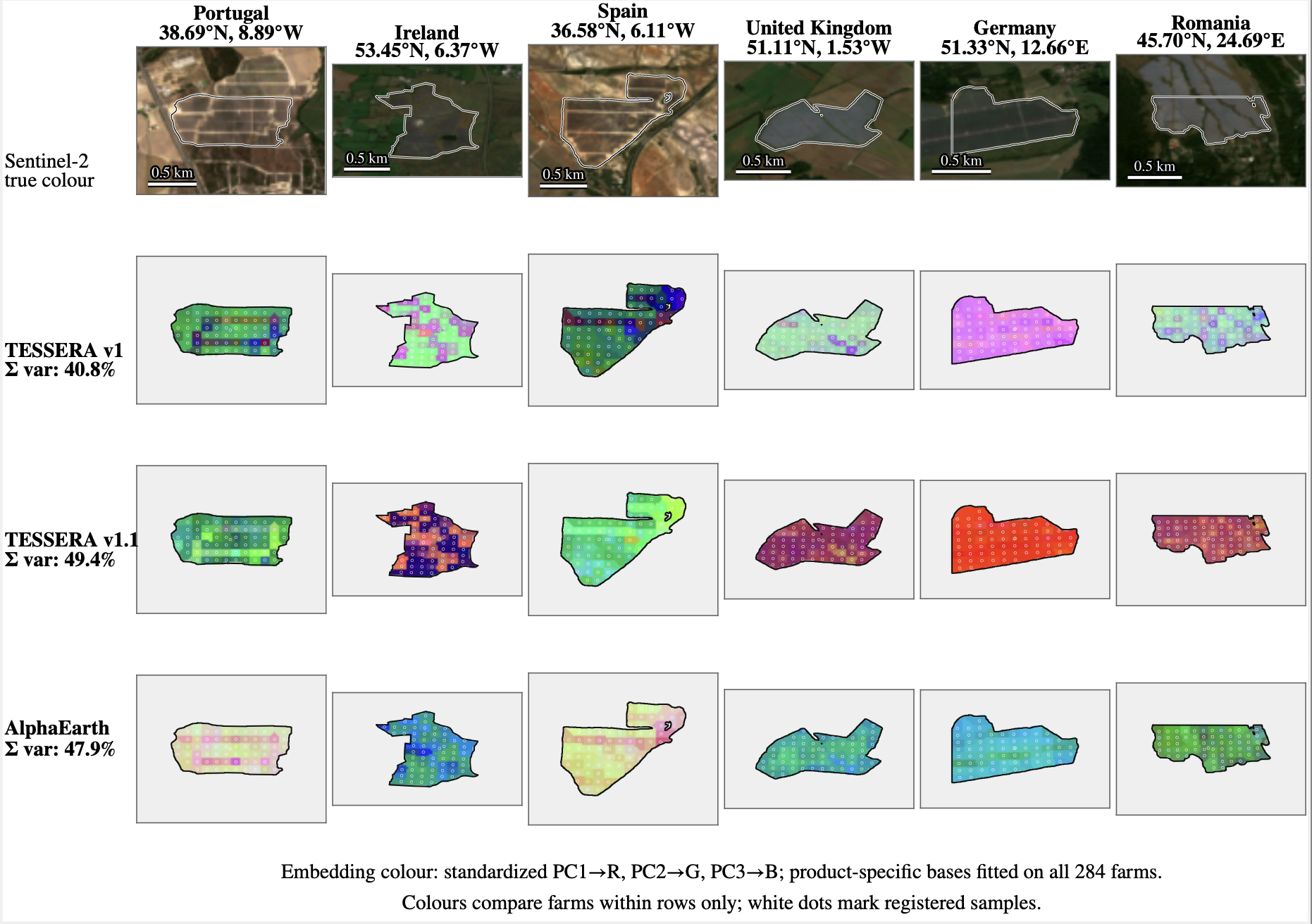}
\caption{Sentinel-2 imagery and PCA-coloured embeddings for six solar farms; colours are comparable only within each rows.}
\label{fig:farm_images}
\end{figure}


\textbf{Dataset and spatial design.} We chose solar farms to isolate geographic effects within a single land-use class with relatively consistent materials and radiometric properties worldwide. We selected 284 solar farms across eight countries from the 2024 Q2 Global Renewables Watch release \cite{robinson2025global}, retaining only farms that passed our own
visual and coverage checks and had valid 2024 coverage in Tessera v1, Tessera v1.1, and AlphaEarth. The final selected solar farms contains 231 farms in Spain, 15 each in France and the United Kingdom, 8 in Germany, 6 in Ireland, 5 in the Netherlands, and 2 each in Portugal and Romania. 
Spain accounted for 81.3\% of the sample, indicating an uneven geographic distribution within the dataset. For each farm, pixel-level embeddings were aggregated using component-wise medians to reduce sensitivity to outliers. Sentinel-2 center pixel and $3\times3$ patch representations served as appearance-based control (Fig.~\ref{fig:farm_images}), providing a comparison with geographic information recoverable from local multispectral features.

\noindent
\textbf{Permutation significance and baseline comparison.} 
Coordinate prediction performance was compared with a training-range uniform random baseline. For each permutation, we reassigned block-centroid locations while preserving within-block coordinate offsets. $B=999$ draws (minimum possible $p=1/1000$), was used to test two null hypotheses: that the embeddings contain no recoverable geographic-coordinate information, and that the geography-embedding distance association is no stronger than under a random relabeling of spatial blocks. For each, the full nested procedure (Ridge fit with inner-fold parameter selection, or the Spearman correlation against permuted geographic distances) was rerun on every permutation $b$, giving a statistic $T_b$ where larger values indicate stronger apparent geographic structure ($T=-E$, the negative held-out geodesic error, for coordinate predicting; $T=|\rho|$, the absolute Spearman coefficient, for the distance association). Both tests reduce to the same empirical $p$-value,
\[
p=\frac{1+\sum_{b=1}^{B}\mathbf{1}\!\left(T_b\geq T_{\mathrm{obs}}\right)}{B+1},
\]
which saturates at $0.001$ once the observed statistic clears all $B$ permutations, establishing that the embeddings contain geographic information beyond the marginal spatial layout of blocks and farms preserved under permutation.

\noindent
\textbf{Farm location prediction and predictor comparison.} We evaluated location prediction using ordinary linear regression, ridge regression, and multilayer perceptrons (MLPs) with $2$-, $4$-, $6$-, and $8$- hidden layers.
To reduce short-range spatial dependence between training and test
samples, farms within 25 km were grouped into dependency components
that could not be split across folds. 

We used the same five-fold spatial cross-validation across all the prediction models and embedding representations: each fold served once as the test set, with the remaining folds used for training. For ridge regression and MLPs, we used regularization to penalize large model weights and limit overfitting. The penalty strength is denoted by $\lambda$ for ridge regression and controlled by the weight-decay parameter for MLPs. For each model run, we selected this parameter using the training data (four-fold spatial validation), keeping farms from the same spatial block together. 

In addition, we compared prediction errors against a uniform random baseline that independently guesses latitude and longitude within the ranges observed in the training farms. Prediction accuracy was measured by the geodesic distance between predicted and actual farm coordinates. We estimated 95\% confidence intervals by resampling the spatial blocks and compared prediction for the same test farms across different models. 

To test how prediction depends on nearby training examples, we varied the training exclusion distance from 25 to 100 km in 5 km increments. For each test fold and distance, we removed any training farm whose centroid lies within that distance of any test farm. Test farms and fold assignments remain fixed throughout the sweep. All models and representations used identical training and test sets at each distance. The 25 km setting reproduces the original training split.

Finally, we conducted leave-one-country-out evaluation using ordinary linear regression, ridge regression, and $8-$layer MLPs to predict the farm coordinates when an entire country was excluded from training.


\begin{table*}[!t]
\centering
\caption{Primary embedding distance-geographic distance correlation and coordinate-decoding results. Brackets give 95\% spatial-block-bootstrap confidence intervals.}
\label{tab:primary}
\footnotesize
\resizebox{0.86\textwidth}{!}{%
\begin{tabular}{lccccc}
\toprule
Input & Dim. & Spearman $\rho$ [95\% CI] & Mean error, km [95\% CI] & Median error, km [95\% CI] & Equal-country mean, km [95\% CI] \\
\midrule
AlphaEarth & 64 & 0.751 [0.664, 0.808] & 179.7 [153.9, 231.0] & 144.5 [127.6, 188.2] & 333.3 [201.3, 486.0] \\
Tessera v1.1 & 128 & 0.424 [0.344, 0.514] & 235.9 [192.3, 331.3] & 174.9 [150.3, 235.0] & 613.9 [366.6, 923.7] \\
Tessera v1 & 128 & 0.386 [0.308, 0.492] & 256.7 [207.0, 361.9] & 178.0 [152.5, 250.3] & 676.1 [432.1, 940.8] \\
Sentinel-2 center-pixel & 13 & 0.294[0.179,0.443] & 391.9[310.2, 518.9] & 304.6 [236.5, 407.5] & 815.2 [490.4, 1169.0] \\
Sentinel-2 $3\times3$ patch & 117 & 0.288 [0.178, 0.396] & 380.8 [303.9, 502.3] & 289.8 [245.1, 390.4] & 807.4 [466.3, 1178.1] \\
\bottomrule
\end{tabular}}
\end{table*}

\begin{figure*}[!t]
\centering
\includegraphics[width=0.75\linewidth]{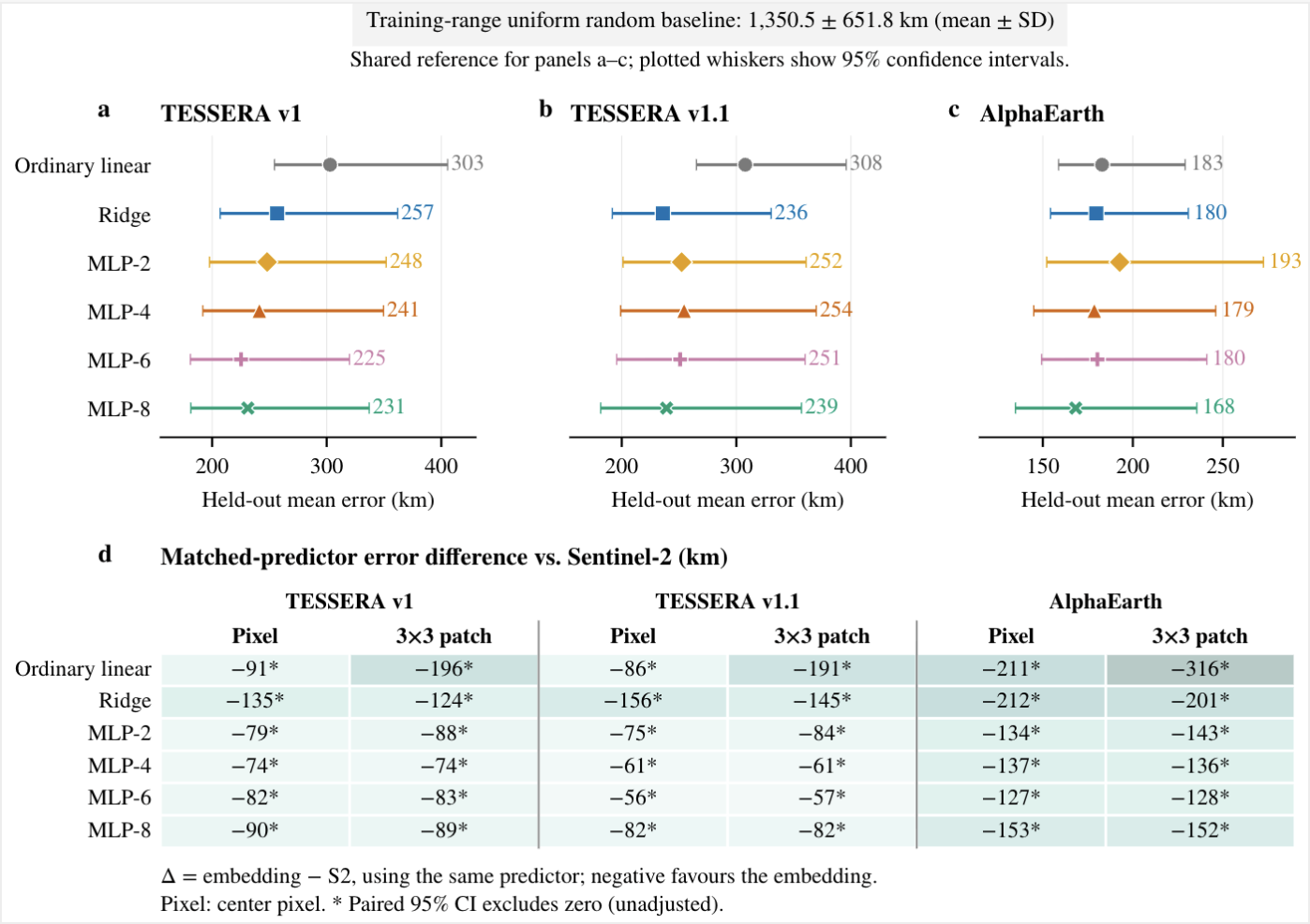}
\caption{Farm location prediction under spatial cross-validation. (a–c) Mean geodesic errors for three embedding products and six predictors. (d) Paired error differences relative to Sentinel-2 controls using the same predictor.}
\label{fig:models}
\end{figure*}

\section{Results}

\subsection{Embedding distance follows physical distance}

All three EO foundation model representations showed positive distance-geography associations (Table~\ref{tab:primary}). Spearman $\rho$ was 0.386 [0.308, 0.492] for Tessera v1, 0.424 [0.344, 0.514] for Tessera v1.1, and 0.751 [0.664, 0.808] for AlphaEarth. The Sentinel-2 center-pixel and 3 × 3 patch controls yielded Spearman correlations of 0.294 [0.179, 0.443] and 0.288 [0.178, 0.396], respectively.  
Permutation tests for Spearman correlation and Ridge regression yielded $p=0.001$, providing strong evidence of existing location predictability, and an association between embedding cosine distance and geodesic distance under the specified permutation tests. The Tessera embeddings showed lower correlations than AlphaEarth and higher point estimates than the Sentinel-2 controls.


\subsection{Farm location prediction from embeddings}

\noindent
\textbf{Geographic location prediction across embeddings.} 
AlphaEarth yielded lower mean location-prediction errors than either TESSERA representation for every predictor tested (168–193 km versus 225–308 km) (Fig.~\ref{fig:models}a-c). TESSERA v1.1 did not consistently improve upon v1: it performed better with ridge regression but had higher mean errors with ordinary linear regression and all MLPs. All three representations substantially outperformed the uniform random baseline (1,350.5 km), indicating that they retained geographic location information useful for predicting held-out farm locations. They also yielded lower mean prediction errors than both Sentinel-2 baselines, center pixel and $3 \times 3$ patch (Fig.~\ref{fig:models}d).  

\noindent
\textbf{Effect of prediction model choice and MLP depth.} 
For each embedding representation, increasing the complexity of prediction models did not consistently reduce error. Ridge yielded lower mean errors than ordinary linear regression for both TESSERA versions and a similar mean error for AlphaEarth(Fig.~\ref{fig:models}a-c). They also yielded lower mean prediction errors than both Sentinel-2 baselines, center pixel and 3×3 patch (Fig.~\ref{fig:models}a-c).

\begin{figure}[t]
\centering
\includegraphics[width=\linewidth]{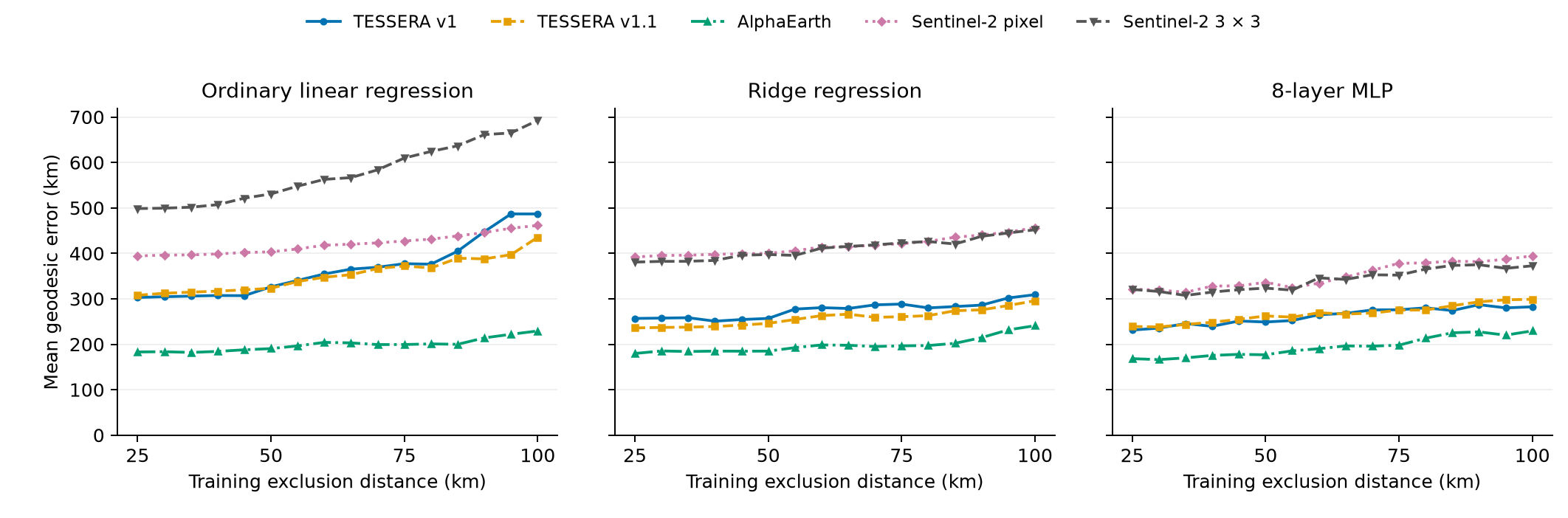}
\caption{Sensitivity of farm location prediction to spatial separation. Mean geodesic error with equal farm weighting as the training exclusion distance increases from 25 to 100 km, using ordinary linear regression, ridge regression, and an 8-layer MLP. Test farms and folds remain fixed across distances.}
\label{fig:distance-dependence}
\end{figure}

\subsection{Location prediction becomes less accurate as spatial separation increases}

We assessed sensitivity to spatial separation by increasing the training exclusion distance from 25 to 100 km in 5 km increments while keeping the test farms and folds fixed. We evaluated ordinary linear regression, ridge regression, and an $8-$layer MLP using the same evaluation procedure and grouped inner-fold hyperparameter selection as in the preceding coordinate-prediction experiment.

Prediction errors generally increased with the exclusion distance, although the magnitude depended on the prediction model and representation. Ordinary linear regression showed larger increases for the TESSERA embeddings and Sentinel-2 $3\times3$ features, whereas ridge regression and the 8-layer MLP showed more gradual increases. These results suggest that its relative advantage persists when nearby training farms are excluded, while absolute prediction accuracy remains sensitive to the evaluation design.


\subsection{Location-prediction performance varies across held-out countries.}

\begin{figure}[!t]
\centering
\includegraphics[width=0.95\linewidth]{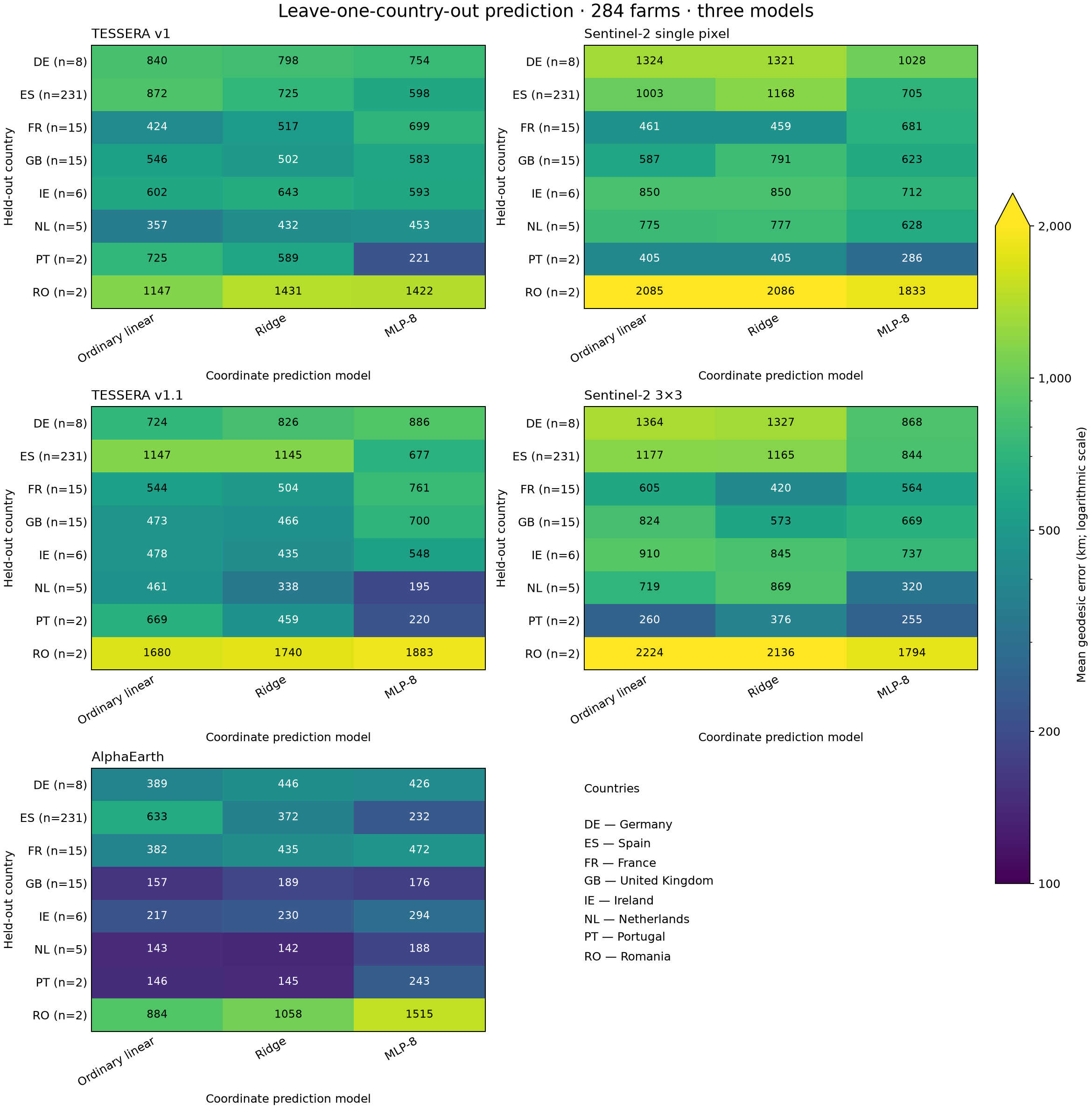}
\caption{Leave-one-country-out farm location prediction. Mean geodesic errors for three predictors and five input representations, with each country excluded in turn from predictor training.}
\label{fig:country-heald-out}
\end{figure}

To test whether geographic information learned from the sampled regions transfers to countries excluded from predictor training, we perform leave-one-country-out evaluation as a geographic generalization test (Fig.~\ref{fig:country-heald-out}). For each country, all farms from that country are held out for testing, while the prediction model is trained on the remaining countries using the same preprocessing and grouped inner-fold hyperparameter selection as the previous coordinate prediction experiment. 

Across the three prediction models, AlphaEarth yields errors of 157--189 km for the United Kingdom and 217--294 km for Ireland, compared with 466--700 km and 435--643 km for the two TESSERA versions, respectively. For Spain, which contains 231 of the 284 farms, AlphaEarth errors range from 232 to 633 km, compared with 598--1,147 km for TESSERA and 705--1,177 km for Sentinel-2.

The relative performance of the predictors also depends on the held-out country. With AlphaEarth, MLP-8 yields a lower error for Spain than ordinary linear regression or ridge regression (232 km versus 633 and 372 km), but a higher error for Romania (1,515 km versus 884 and 1,058 km). Romania has the largest error within every representation–predictor combination, with values ranging from 884 to 2,224 km. AlphaEarth does not yield the lowest error in every comparison: for Portugal, MLP-8 errors are 220--221 km with TESSERA and 243 km with AlphaEarth. These results indicate variation in geographic transfer across the evaluated countries, with errors remaining on the order of hundreds of kilometres in many settings. Country-level comparisons should be interpreted cautiously because the held-out samples are uneven, including only two farms each for Portugal and Romania. 

\section{Discussion and Conclusion}

Using a solar farm dataset restricted to one semantic class, we show that embeddings from all three EO foundation models contain recoverable geographic information. AlphaEarth shows the strongest geographic distance association and lowest location-prediction error across the tested prediction models, while Tessera v1.1 is comparable to v1. Prediction errors generally increase as nearby training farms are excluded, and performance varies across held-out countries.

The generalizability of these findings is limited by uneven geographic coverage, with 81.3\% of farms in Spain. Greater spatial separation and country holdouts also reduce training data, complicating the interpretation of prediction errors. The ability to predict locations may partly reflect country-specific or dataset-specific patterns rather than fine-grained geographic information. We therefore do not claim that coordinates are explicitly encoded, and the desirability of geographic dependency ultimately depends on the downstream task, being useful for location-sensitive applications such as climate mapping but potentially undesirable for global similarity search. We recommend assessing geographic information in the embeddings alongside downstream performance and interpreting it in relation to the task’s need for location information.

\raggedbottom
\clearpage

\end{document}